\documentclass{article}
\usepackage{spconf,amsmath,epsfig}
\usepackage{times}
\usepackage{epsfig}
\usepackage{graphicx}
\usepackage{amsmath}
\usepackage{amssymb}
\usepackage[export]{adjustbox}
\usepackage{subcaption}
\usepackage{graphbox}

\usepackage{mathtools}
\mathtoolsset{showonlyrefs} 
\usepackage{siunitx}

\usepackage[pagebackref=true,breaklinks=true,letterpaper=true,colorlinks,bookmarks=false]{hyperref}
\DeclareGraphicsExtensions{.pdf,.jpg,.png}
\graphicspath{{figs/}}

\newcommand{\secref}[1]{Section~\ref{sec:#1}}
\newcommand{\figref}[1]{Figure~\ref{fig:#1}}

\let\origfootnote\footnote
\renewcommand{\footnote}[1]{\kern.06em\origfootnote{#1}}
\newcommand{\punctfootnote}[1]{\kern-.06em\origfootnote{#1}}

\title{Learning to Map Nearly Anything}
\name{Tawfiq Salem, Connor Greenwell, Hunter Blanton, Nathan Jacobs}
\address{Department of Computer Science, University of Kentucky, USA. \\\small{(salem,connor,hunter,jacobs)@cs.uky.edu}}

\begin{document}
\maketitle
\begin{abstract}
Looking at the world from above, it is possible to estimate many properties of a given location, including the type of 
land cover and the expected land use. Historically, such tasks have relied on relatively coarse-grained categories due 
to the difficulty of obtaining fine-grained annotations. In this work, we propose an easily extensible approach that makes 
it possible to estimate fine-grained properties from overhead imagery. In particular, we propose a cross-modal distillation 
strategy to learn to predict the distribution of fine-grained properties from overhead imagery, without requiring 
any manual annotation of overhead imagery. We show that our learned models can be used directly for applications 
in mapping and image localization.
\end{abstract}


\section{Introduction}
\label{sec:intro}
Traditional approaches to pixel-level labeling of remote sensed imagery rely on the manual specification 
of semantic categories.  In our view, this limits the ability to predict categories for which a human 
annotator has low confidence.  This means that we are unable to learn to make predictions about less 
certain categories.  We propose to overcome this problem using a weakly-supervised learning strategy 
that uses manually specified labels in a domain for which humans are confident (ground-level imagery) 
but allows us to make less confident predictions for overhead imagery.  With this strategy we are able 
to estimate probability distributions over categories that would normally be considered too difficult 
for overhead imagery understanding, due to the lack of available training data.
\begin{figure}
    \centering
    \includegraphics[scale=0.4, width=.99\linewidth]{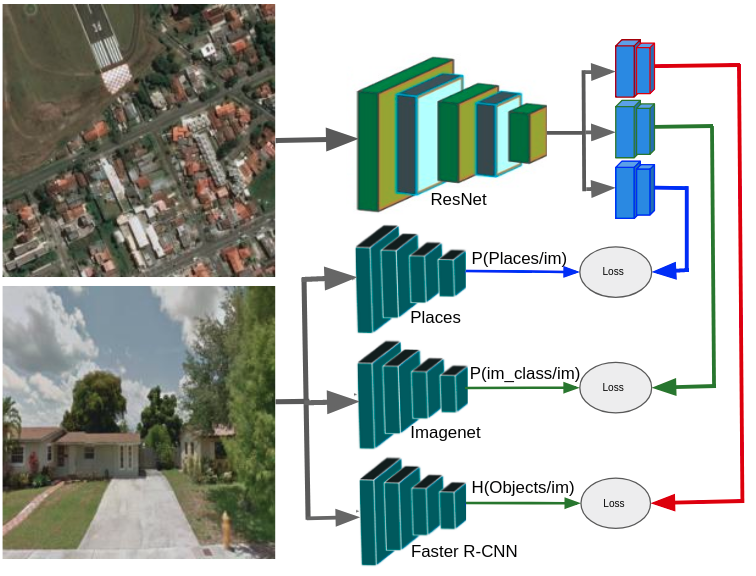}
    \caption{An overview of our network architecture.}
    \label{fig:architecture}
\end{figure}

In particular, using large numbers of GPS-tagged consumer photographs, we use off-the-shelf networks 
for image classification, scene classification, and object detection in ground-level images to build 
a sparse training dataset for overhead image understanding. We extend the approach in~\cite{workman2015wide} 
by modeling the distribution of labels. The previous work assumed that the predictions of the ground-level 
images were distributed in a Gaussian manner, and only learned to predict the mean of the distribution. 
This is problematic when a particular overhead image could have multiple possible interpretations from 
a ground-level perspective.  For example, if the overhead image contains a beach and a parking lot, then 
the ground-level image may be of a beach or a parking lot, depending on the orientation.
To capture such uncertainty, we model the distribution of ground-level image labels as samples from 
a Dirichlet distribution. We use a multi-task approach and predict the parameters of prior distributions 
over three label spaces: scene categorization~\cite{zhou2017places}, image
classification~\cite{krizhevsky2012imagenet}, and object detection~\cite{ren2015faster}. 
    
\begin{figure*}
  \centering
  \begin{subfigure}{0.49\linewidth}
       \raisebox{-0.5\height}
       {\includegraphics[height=1.73in,width=.40\linewidth]{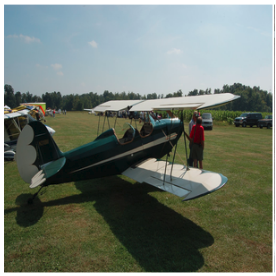}}
       \raisebox{-0.5\height}{\includegraphics[width=.5\linewidth]{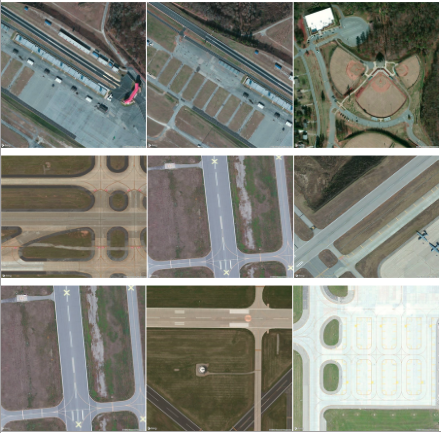}}
          \caption{}
       \label{fig:airport}
 \end{subfigure}
 \begin{subfigure}{0.49\linewidth}
    \raisebox{-0.5\height}
       {\includegraphics[height=1.68in, width=.4\linewidth]{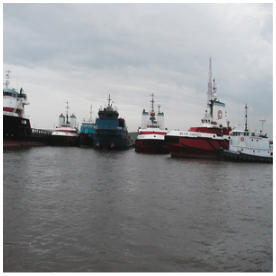}}
       \raisebox{-0.5\height}{\includegraphics[width=.5\linewidth]{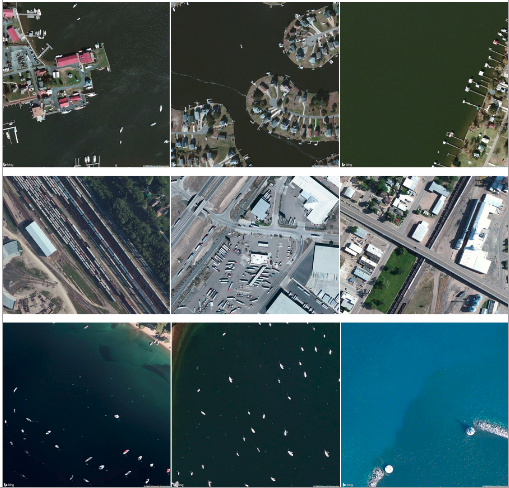}}
       \caption{}
       \label{fig:boats}
 \end{subfigure}
  \caption{For a given ground image, we show the top-$3$ overhead images that give the highest probability 
  for the given image. The top row is based on Places, the second on ImageNet, and the last on object counts.}
  \label{fig:matching}
\end{figure*}

\section{Related Work}
Many recent works jointly reason about ground-level and overhead image viewpoints. 
Zhai et al.~\cite{zhai2017crossview} incorporate a transformation between co-located ground-level and 
overhead images to learn semantic features for overhead imagery. Cross-view image geolocalization
strategies~\cite{workman2015wide,lin2013cross,Hu_2018_CVPR,lin2015learning}
learn a feature mapping between the two viewpoints in order to leverage densely available overhead imagery. 
Other works have used large-scale image collections to map properties of the world. 
For example, Lee et al.~\cite{lee2015predicting} estimate geo-informative attributes such as population 
density and demographic properties. Another work by Salem et al.~\cite{salem2018soundscape} proposed 
an approach for constructing an aural atlas, which captures the geospatial distribution of soundscapes. 
Most similar to our work, Workman et al.~\cite{workman2015wide} proposed an approach for cross-view 
training to learn similar feature representation for co-located ground and overhead images and use this for 
geolocalization. Greenwell et al.~\cite{greenwell2018goes} proposed a similar cross-view learning approach to learn a model that is capable of predicting the type and count of objects that are likely to be seen from a ground-level perspective conditioned on the overhead image. The previous two methods work on a single ground-level distribution. We propose a general architecture that can learn all labels that we can get from the ground perspective.  

\section{Approach}
We propose a cross-view training strategy (\figref{architecture}) that uses pre-existing CNNs to extract categorical
distributions of ground-level images to provide a weak signal for predicting the parameters of probabilistic models
conditioned on co-located overhead imagery. We simultaneously learn three such probabilistic models that model separate
ground-level distributions.

\subsection{Dataset}
\label{sec:dataset}
In our work, we use the \num{551851} Flickr geotagged ground-level images, and corresponding overhead images, contained in the Cross-View USA (CVUSA)
dataset~\cite{workman2015wide}. For each ground-level image, we extracted two categorical distributions: the first 
over \num{365} categories using a VGG16-Places365 scene recognition model trained on Places2~\cite{zhou2017places}, and
the second over \num{1000} categories using VGG16-ImageNet trained for the task of image
classification~\cite{krizhevsky2012imagenet}. We also use the histogram of object counts in
each image.  Following the approach of Greenwell et al.~\cite{greenwell2018goes}, we use the Faster R-CNN ResNet
101~\cite{ren2015faster} detector trained on the MS-COCO challenge dataset~\cite{lin2014microsoft}. We split this dataset into $93\%$ training, $2\%$ validation, and $5\%$ testing subsets.
    
\subsection{Distribution Representation} 
We use two common distributions to model priors over ground-level image features: 
Dirichlet and Poisson. The Dirichlet distribution is the conjugate prior of the categorical distribution, 
meaning samples drawn from a Dirichlet distribution are themselves categorical distributions. 
Given parameters $\alpha_i$, the probability density function is given by the following equation:   
\begin{equation}\label{eq:2}
    f(x_1,...,x_k;\alpha_1,...,\alpha_k) = \frac{1}{B(\alpha)}\prod_{i=1}^kx_i^{\alpha_i-1}
\end{equation}
where $B(\alpha)$ is a normalizing constant. Using this we can model the one-to-many relationship between overhead imagery and potential ground-level scene and object probabilities through a discrete set of parameters. The Poisson distributions 
describe the likelihood of an event happening $k$ times in some fixed interval. In our case, this will be the probability of $k$ objects of a class being present in the spatial extent of the scene. For each object class, the probability of $k$ objects 
of that class appearing is given by the following equation:  
\begin{equation}\label{eq:2}
    P(k) = e^{-\lambda}\frac{\lambda^k}{k!}
\end{equation}
where $\lambda$ is the interval rate, which varies per class.
Using a Poisson distribution, we can directly model probabilities of not 
only the types of objects expected in a ground-level scene, but also the number of expected occurrences.

\subsection{Network Architecture}
\label{sec:arch}
Our network (\figref{architecture}) has two main components. The first is a collection of 
pre-trained models that we use for extracting ground-level predictions. 
The second is a shared CNN which takes an overhead image as input and produces a feature 
which is passed to $3$ separate prediction heads. These heads separately predict 1) 
parameters of a Dirichlet distribution over scene categories, 2) parameters of 
a Dirichlet distribution over ImageNet classes, and 3) parameters of Poisson distributions
over the histogram of objects in the image for each overhead image. Each head consists of 
two fully connected layers. The first layer of each head contains $1024$ neurons. 
The second layer is different for each task, $365$, \num{1000}, and $91$ respectively. 
During training we use three losses, one for each distribution, that minimize the mean 
negative log-likelihood of the resulting probability distributions as in the following equation:
\begin{equation}
    L = min\ \frac{1}{N}\sum_a -log(p(g|a,w))\label{eq:3}
\end{equation}  
\noindent where $g$ represents the features (categorical distributions, object counts) coming 
from the ground-level image, $a$ is the overhead imagery, and $w$ represents the learned weights . 

\subsection{Implementation Details}
Our model is implemented in TensorFlow. We initialize ResNet-v2-50~\cite{he2016identity} by pre-training it to predict two distributions over image-level scene and object categories from overhead imagery using the KL-divergence as a loss function. The trained ResNet is then frozen for subsequent fine-tuning of three prediction heads. The heads are initialized randomly using Xavier initialization. We optimize each head simultaneously by minimizing the negative log-likelihood as explained in \secref{arch}. For both ResNet pre-training and training of the prediction heads, we use the Adam optimizer with a learning rate of $0.001$ and a weight decay factor of $0.0005$ for $6$ epochs with batch size $32$. The input images are re-sized to $224 \times 224$, scaled to $[\num{-1},1]$, and augmented by random horizontal and vertical flipping.

\begin{figure}[t!]
\centering
 {\includegraphics[width=.98\linewidth]{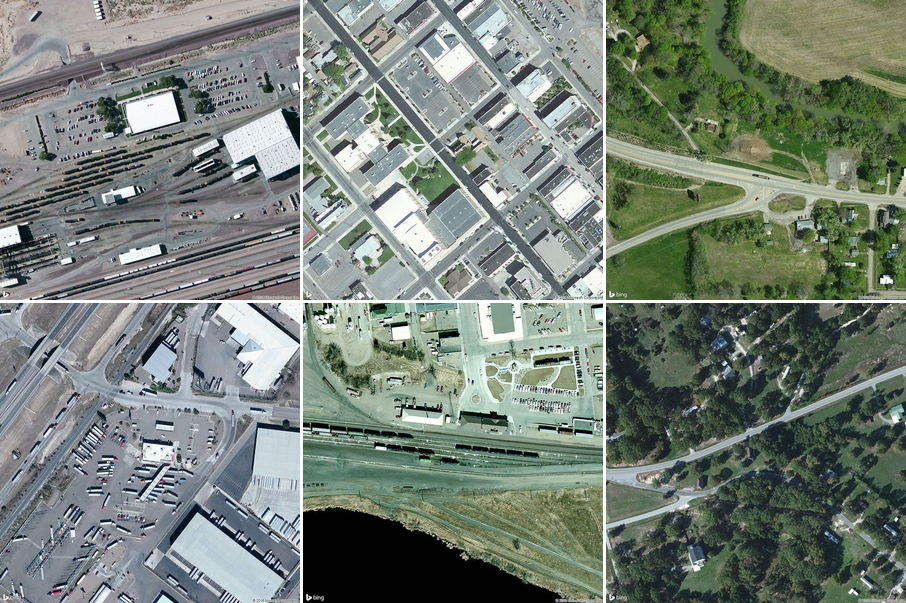}}
\caption{Overhead images with the highest scores for the {\em car} label. The {\em park} label score is increased from left to right, transitioning the images from industrial to rural scenes while focusing on roads. Each column represents multiple images with similar scores for the query labels. }
\label{fig:cross_labels}
\end{figure}
    
\section{Evaluation}
After we trained our proposed model using the dataset defined in \secref{dataset}, we conducted several experiments that highlight potential applications of our model.

\subsection{Cross-View Image Retrieval}
Using the test-set defined in \secref{dataset}, we extract the output parameters that define the three distributions for each overhead image. The distributions are used to compute the 
log-probability for any query ground-level image to identify the top-3 overhead images with highest probability. Two qualitative examples are shown in \figref{matching}. The left image shows the query ground-level image and on the right we show the top-3 overhead images for each distribution (from top to bottom row: Places, ImageNet, and object counts). For example, in (a) the top row shows the top-3 matching overhead images that seem to contain highways and baseball field where in the middle and bottom contain runways. 

\begin{figure}[t!]
\centering
\raisebox{-0.5\height}{\includegraphics[height=2.5cm,width=.35\linewidth]{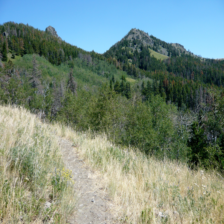}}
\raisebox{-0.5\height}{\includegraphics[height=2.5cm,width=.60\linewidth]{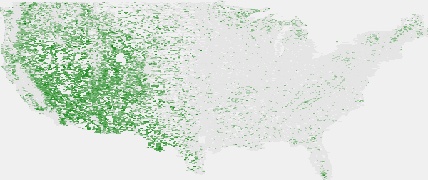}}
\raisebox{-0.5\height}{\includegraphics[height=2.5cm,width=.35\linewidth]{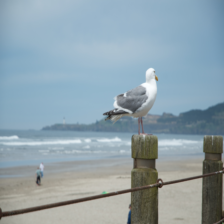}}
\raisebox{-0.5\height}{\includegraphics[height=2.5cm,width=.60\linewidth]{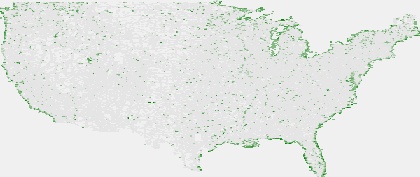}}
\raisebox{-0.5\height}{\includegraphics[height=2.5cm,width=.35\linewidth]{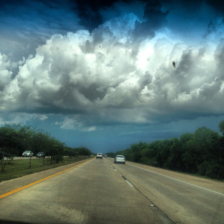}}
\raisebox{-0.5\height}{\includegraphics[height=2.5cm,width=.60\linewidth]{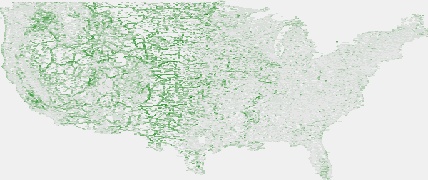}}
\caption{Given a query ground-level image (left), we can construct a heatmap (right) that represents 
the score where the greener the dot on the map the more likely the image was taken in that location.}
\label{fig:localization} 
\end{figure}

\subsection{Location Search Using Joint Attributes}
Using our trained model, we can identify locations that contain specific types of scenes, objects, or some combination of the two. For example, in \figref{cross_labels}, images with high scores for {\em car} (an ImageNet label) are shown. The images are further sorted from left to right by increasing score for {\em park} (a Places label). This results in a set of images centered around roads, with a smooth transition from industrial to rural as the {\em park} score increases. This highlights the ease of specifying location-search query parameters using the fine-grained semantics available in our model.

\subsection{Cross-View Localization}
In this experiment, we constructed a reference database by taking all the overhead images in our dataset (training and testing), and for each we predict the parameters for the three different distributions. Then for every ground-level image in the test-set, we can get a score for every overhead image in the reference database. In~\figref{localization}, we show some qualitative results based on the Places-Dirichlet distribution. The heatmap (right) represents the likelihood that an image was captured at a specific location, where the greener the dot, the more likely the given ground-level image (left) was taken at that location. In the middle row, our method clearly identifies the query image as having been captured along the coast of USA. Similarly, we can generate the heatmap based on the other two distributions.

\begin{figure}[t!]
\centering
\includegraphics[scale=0.4, width=.99\linewidth]{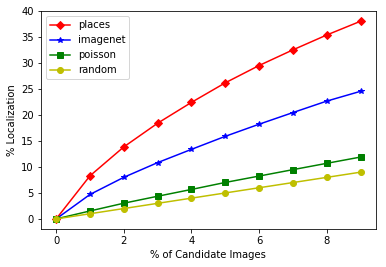}
\caption{Localization accuracy of the different learned probabilistic models on the test-set of the ground-level
imagery }
\label{fig:accuracy}
\end{figure}

We also quantitatively evaluated the model for this task by calculating the localization accuracy on the test-set. In \figref{accuracy}, we show the percent of the correctly localized query ground-level images (y-axis) at various error thresholds (x-axis), percentages of overhead images ranked higher than the true overhead image. The localization accuracy of the different learned probabilistic models show that using the distribution based on Places gives the highest accuracy followed by ImageNet, and the worst performance (but still better than random) is the Poisson distribution defined based on the object counts. We suspect the low performance of the Poisson is because most of the \num{91} object categories are not geographically discriminative attributes and some of these object categories are difficult to learn from the overhead imagery. 

\section{Conclusion}
We created a location-dependent model for mapping different ground-level properties using overhead imagery. We show how our model can be used to generate maps at varying spatial scales. In the future, we will extend this work to be conditioned on time, because what you can expect to see and experience at a location are highly dependent on when and where the image was captured.\\

\noindent\textbf{Acknowledgments} 
We gratefully acknowledge the support of NSF CAREER award (IIS-1553116).

{\small
\bibliographystyle{IEEEbib}
\bibliography{biblio}
}
\end{document}